\documentclass[sigconf, 9pt, nonacm]{acmart}

\usepackage{threeparttable}
\usepackage{amsthm}
\usepackage{textcomp}
\usepackage{booktabs}
\usepackage{bm}
\usepackage{multirow}
\usepackage{algorithm}
\usepackage{epstopdf}
\usepackage{epsfig}
\usepackage{enumitem}
\usepackage{booktabs}
\usepackage{diagbox}
\usepackage{amsmath,amsthm}

\usepackage{multirow}
\usepackage[normalem]{ulem}
\usepackage{makecell}
\usepackage{xcolor}
\usepackage{colortbl}
\usepackage[pscoord]{eso-pic}
\usepackage{xcolor}
\usepackage{stfloats}
\newcommand{\qdel}[1]{}

\newcommand{\Qdel}[1]{}

\newcommand{\tdel}[1]{}

\newcommand{\hdel}[1]{}

\AtBeginDocument{%
  \providecommand\BibTeX{{%
    \normalfont B\kern-0.5em{\scshape i\kern-0.25em b}\kern-0.8em\TeX}}}

\setcopyright{none}
\renewcommand\footnotetextcopyrightpermission[1]{} 
\begin{document}

\title{\textit{VeriDispatcher}: Multi-Model Dispatching through Pre-Inference Difficulty Prediction for RTL Generation Optimization}

\author[Zeng~Wang, Weihua~Xiao, Minghao~Shao, Raghu~Vamshi~Hemadri, Ozgur~Sinanoglu, Muhammad~Shafique, Ramesh~Karri]{%
Zeng~Wang$^\dagger$$^\S$,
Weihua~Xiao$^\dagger$$^\S$,
Minghao~Shao$^\dagger$$^\ddagger$$^\S$,
Raghu~Vamshi~Hemadri$^\dagger$,
Ozgur~Sinanoglu$^\ddagger$,
Muhammad~Shafique$^\ddagger$,
Ramesh~Karri$^\dagger$\\[4pt]
$^\dagger$NYU Tandon School of Engineering, New York, USA\\
$^\ddagger$NYU Abu Dhabi, Abu Dhabi, UAE\\[3pt]
\text{\{zw3464, wx2356, shao.minghao, rh3884, ozgursin, muhammad.shafique, rkarri\}@nyu.edu}
}

\begin{abstract}
\textit{Large Language Model}s (\textit{LLM}s) show strong performance in RTL generation, but different models excel on different tasks because of architecture and training differences. 
Prior work mainly prompts or finetunes a single model. What remains not well studied is how to coordinate multiple different LLMs so they jointly improve RTL quality while also reducing cost, instead of running all models and choosing the best output. 
We define this as the \emph{multi-LLM RTL generation} problem. We propose \textit{\textbf{VeriDispatcher}}, a multi-LLM RTL generation framework that dispatches each RTL task to suitable LLMs based on pre-inference difficulty prediction. For each model, we train a compact classifier over semantic embeddings of task descriptions, using difficulty scores derived from benchmark variants that combine syntax, structural similarity, and functional correctness. 
At inference, VeriDispatcher uses these predictors to route tasks to a selected subset of LLMs. 
Across $10$ diverse LLMs on \textit{RTLLM} and \textit{VerilogEval}, VeriDispatcher achieves up to $18\%$ accuracy improvement on RTLLM using only $40\%$ of commercial calls, and on VerilogEval maintains accuracy while reducing commercial usage by $25\%$, enabling cost-effective, high-quality LLM deployment in hardware design automation.
\end{abstract}


\maketitle
\begingroup
\renewcommand\thefootnote{$\S$}%
\footnotetext{Authors contributed equally to this research.}%
\endgroup
\section{Introduction}

\textit{Large language model}s (\textit{LLM}s) are transforming hardware design automation, enabling RTL generation~\cite{liu2024rtlcoder, lu2024rtllm}, assertion synthesis~\cite{tian2025assertcoder}, testbench creation~\cite{bhandari2024llm}, and security evaluation~\cite{wang2025verileaky, wang2025vericontaminated, hemadri2025veriloc}. While these advances accelerate design cycles and democratize expertise~\cite{ibnat2025trusting}, most research focuses on improving individual models via better data or pipeline~\cite{liu2024rtlcoder, zhao2025mage}, architectures~\cite{zhu2025codev,yang2025haven}, reasoning~\cite{zhu2025codev, yubeaton2025verithoughts, wang2025salad}, or reflection~\cite{cui2024origen}. This overlooks a practical reality: engineering teams often access diverse models with complementary strengths across task complexities. 
What remains underexplored is how to coordinate multiple LLMs to both improve RTL generation quality and reduce cost compared to running all models and picking the best result. 
We call this the \textit{multi-LLM RTL generation problem}.

Real-world deployments combine commercial APIs, open-source models, and fine-tuned variants, each trained on different data distributions, and alignment objectives.
Without a principled way to match tasks to models, teams either guess or run all models, which is inefficient and risks mismatching task characteristics and model strengths.
What is missing is a pre-generation tool that estimates how well each LLM can handle a Verilog task and how difficult that task is for each model. Existing benchmarks~\cite{liu2023verilogeval, lu2024rtllm} provide only post-hoc difficulty labels, and simulation or synthesis feedback~\cite{wolf2013yosys, iverilog} is available after code has been generated. Simple heuristics such as prompt length correlate poorly with true complexity, since short specifications can encode highly intricate behavior. 

\begin{figure}[!t]
    \centering
    \includegraphics[width=1.0\linewidth]{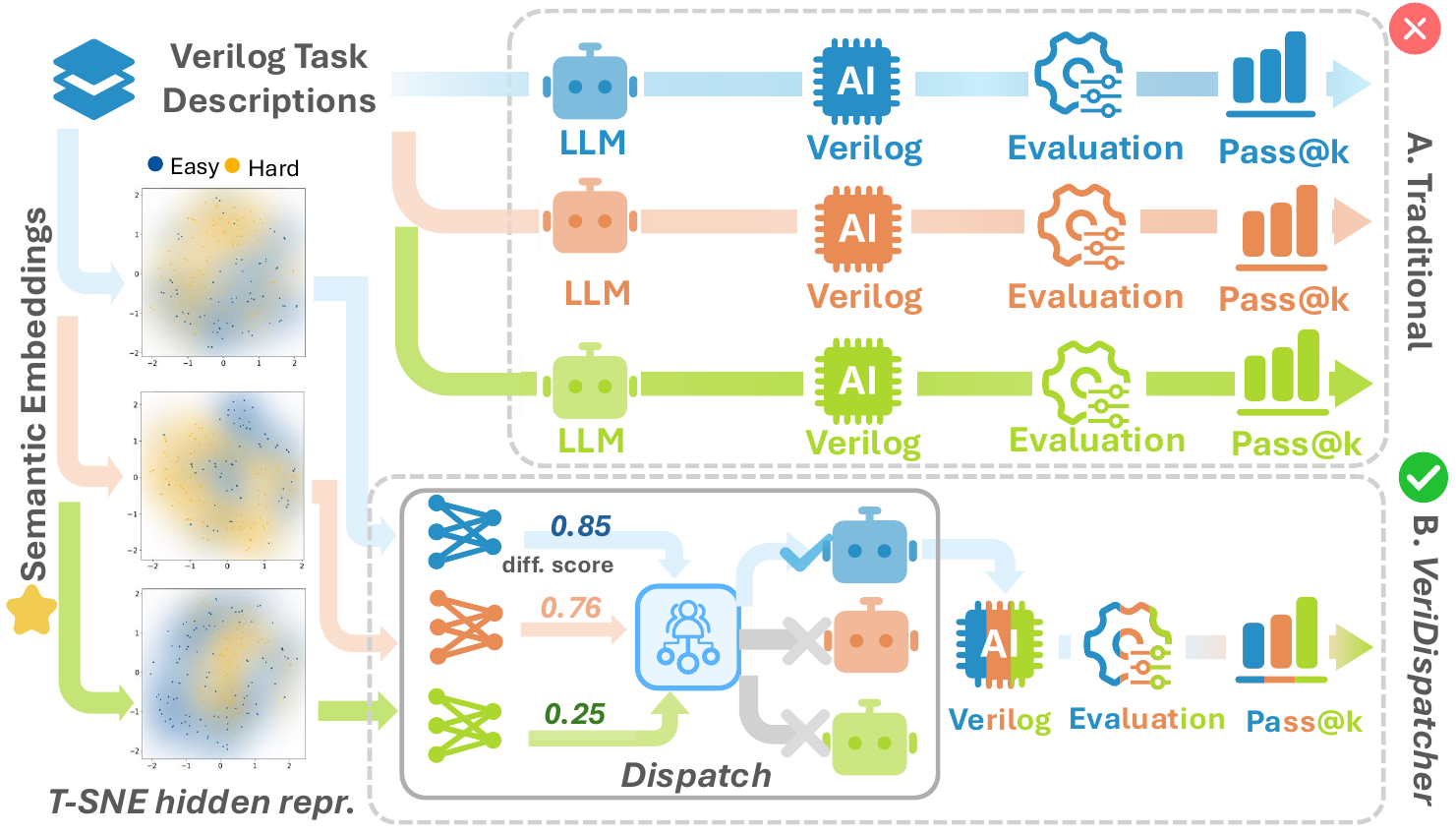} 
    \caption{
    Traditional multi-LLM RTL generation vs.\ difficulty-aware VeriDispatcher.
    }
    \label{fig:motivation}
\end{figure}


To close this gap, we propose \textit{VeriDispatcher}, a framework that improves both generation quality and efficiency by routing each RTL task to the most suitable LLM using \textit{difficulty-aware model selection}. The key idea is that difficulty is model-dependent: a task hard for one model may be easy for another due to differences in training data or architecture. Instead of using a single global predictor, VeriDispatcher trains a lightweight classifier for each model to learn its capability boundary. These classifiers take semantic task embeddings as input and are supervised by empirical generation outcomes, enabling calibrated alignment between tasks and models.

Fig.~\ref{fig:motivation} contrasts the traditional multi-LLM RTL generation framework with our difficulty-aware VeriDispatcher. In the traditional pipeline (top), each RTL task is broadcast to all available LLMs; each model generates RTL, the outputs are evaluated, and the best $\text{pass}@k$ result is selected, achieving good quality but at high cost. On the left, the t-SNE~\cite{maaten2008visualizing} plots visualize semantic embeddings of the same task set for different LLMs, where easy and hard instances cluster differently per model, revealing that task difficulty is inherently model specific. Leveraging this observation, VeriDispatcher (bottom) encodes each task into a semantic embedding and feeds it to a per-model lightweight \textit{Multi-Layer Perceptron} (\textit{MLP}) trained on difficulty scores combining syntax, structural similarity, and functional correctness. At inference time, the predicted difficulty scores route each task only to the LLM expected to handle it best, reducing model calls while maintaining or improving RTL generation quality.

We evaluate VeriDispatcher on \textit{RTLLM}~\cite{lu2024rtllm} and \textit{VerilogEval}~\cite{liu2023verilogeval} across $10$ different LLMs. Results show our approach obtains up to 18.18\% improvement over single-model baselines and random selection dispatching. Improvements persist under confidentiality constraints by dispatching sensitive tasks to on-premise models, ensuring effective model utilization. In summarize, we have the following contributions:
\begin{enumerate}[leftmargin=*, itemsep=0pt, topsep=2pt]
\item We formulate multi-LLM RTL generation as a \textit{difficulty-aware model selection task} that assigns tasks to the most suitable LLM.
\item We present VeriDispatcher, which assigns RTL tasks with model-specific difficulty predictors from semantic embeddings.
\item We conduct a comprehensive evaluation showing strong gains on complex benchmark with detailed insights and analysis.
\end{enumerate}
\section{Background}
\subsection{LLMs for RTL Generation}
LLMs have demonstrated significant potential across hardware design domains~\cite{wang2024llms,shao2024survey}, including RTL code generation~\cite{thakur2023autochip}, assertion synthesis~\cite{kande2023llm}, and testbench development~\cite{bhandari2024llm}. Recent advances have pursued diverse strategies to improve generation quality.
Model architecture innovations include RTL-Coder~\cite{liu2024rtlcoder}, which synthesizes instruction-code pairs to surpass GPT-3.5 performance, and CodeV~\cite{zhao2025codev}, which employs multi-level summarization and FIM-Tag fine-tuning for Verilog and Chisel generation. CodeV-R1~\cite{zhu2025codev} further integrates reasoning distillation with reinforcement learning to achieve state-of-the-art results. OriGen~\cite{cui2024origen} advances open-source RTL generation through code-to-code augmentation and compiler-guided reflection, matching commercial model performance. Novel strategies include HAVEN~\cite{yang2025haven}, which uses non-textual prompting representations, and VeriThoughts~\cite{yubeaton2025verithoughts}, which introduces reasoning-augmented datasets with formal verification labels.
Despite advances in single LLMs and fine-tuned variants for RTL generation, it remains unclear how to intelligently select, from a pool of different LLMs, the model best suited for a given RTL task in a cost-effective way, motivating our focus on pre-inference difficulty estimation from internal embedding representations.

\begin{figure*}[!t]
    \centering
    \includegraphics[width=1\linewidth]{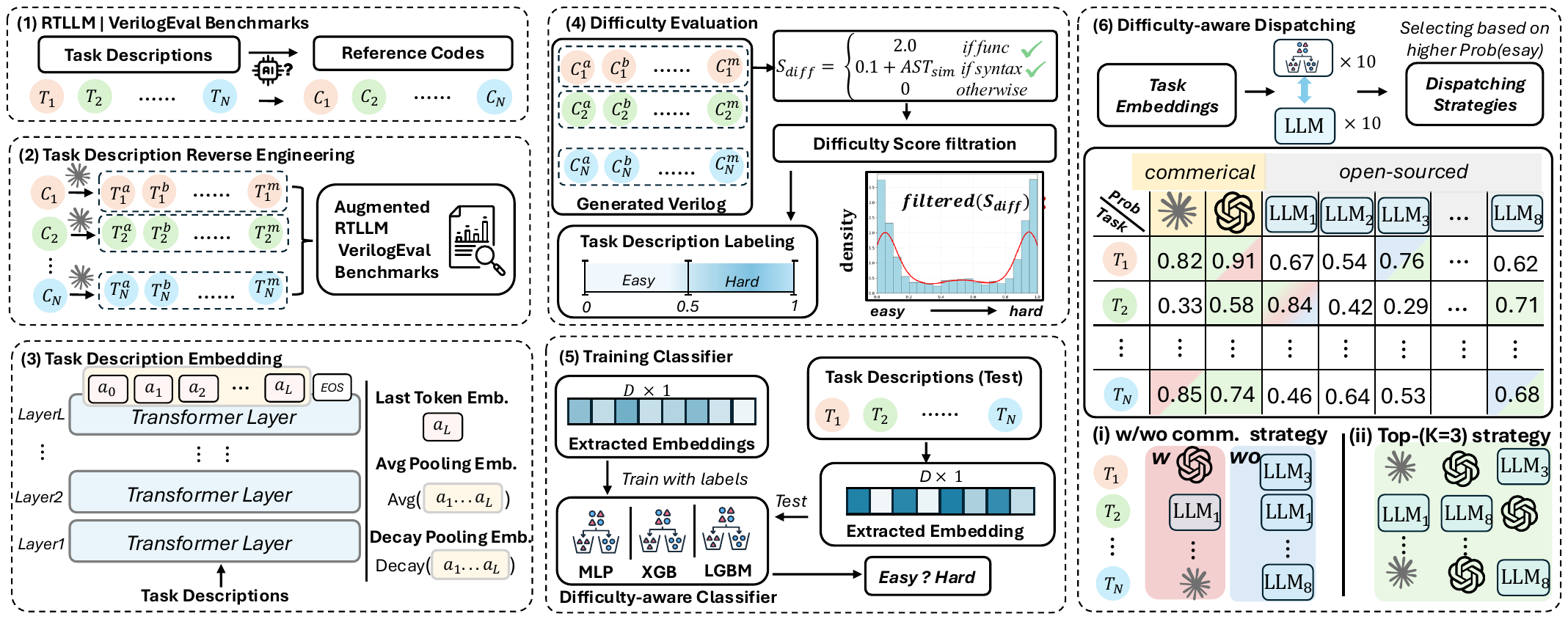} 
    \caption{Complete VeriDispatcher pipeline from task embedding extraction through difficulty-aware model dispatching.}
    \label{fig:veridispatcher}
    \vspace{-2mm}
\end{figure*}
\subsection{Representations and Embeddings of Inputs}
Transformer-based LLMs encode input semantics through latent embeddings and hidden representations~\cite{zhu2025llm, ni2025towards} that correlate with task complexity, reasoning trajectories, and model confidence before generation. Embedding-space analyses in natural language reasoning reveal separable patterns between easy and hard questions, indicating that hidden states inherently capture perceived task difficulty. This enables assessing complexity directly from internal embeddings without token generation.

\subsection{LLMs Difficulty Evaluation}
Existing Verilog benchmarks, VerilogEval~\cite{liu2023verilogeval} and RTLLM~\cite{lu2024rtllm}, advance LLM evaluation but face critical limitations. They rely on subjectively defined difficulty hierarchies that may not reflect model-intrinsic capabilities, and require expensive simulation-based validation through toolchains like Yosys~\cite{wolf2013yosys} and Icarus Verilog~\cite{iverilog}, creating scalability bottlenecks.
Difficulty estimation in other domains employs: (1) statistical methods analyzing performance patterns~\cite{wang2022self}; (2) LLM-as-a-judge frameworks~\cite{cheng2025think, chen2024magicore};  and (3) representation -based methods using hidden embeddings~\cite{zhu2025llm}. While the first two require repeated generation, representation-based approaches enable direct complexity estimation from model-internal states without costly sampling.
However, such methods remain unexplored in hardware design, where syntactic correctness and functional equivalence create unique challenges. This gap prevents scalable evaluation, early risk assessment, and adaptive resource allocation in LLM-driven EDA workflows. Our work addresses this through a representation-based framework that estimates task difficulty directly from hidden states, enabling intelligent resource allocation and security-aware Verilog generation.
\section{Framework}
VeriDispatcher is a lightweight framework for solving multi-LLM available RTL generation:
it routes each task to the most suitable LLM using a pre-inference difficulty predictor that estimates how likely each model is to solve the current task correctly.
As shown in Fig.~\ref{fig:dispatching}, our framework uses semantic embeddings of task descriptions to predict whether a model-specific generation can complete the task before code synthesis, obviating the costly post-generation verification and validation to determine model suitability.

\begin{figure*}[!t]
    \centering
    \includegraphics[width=1.0\linewidth]{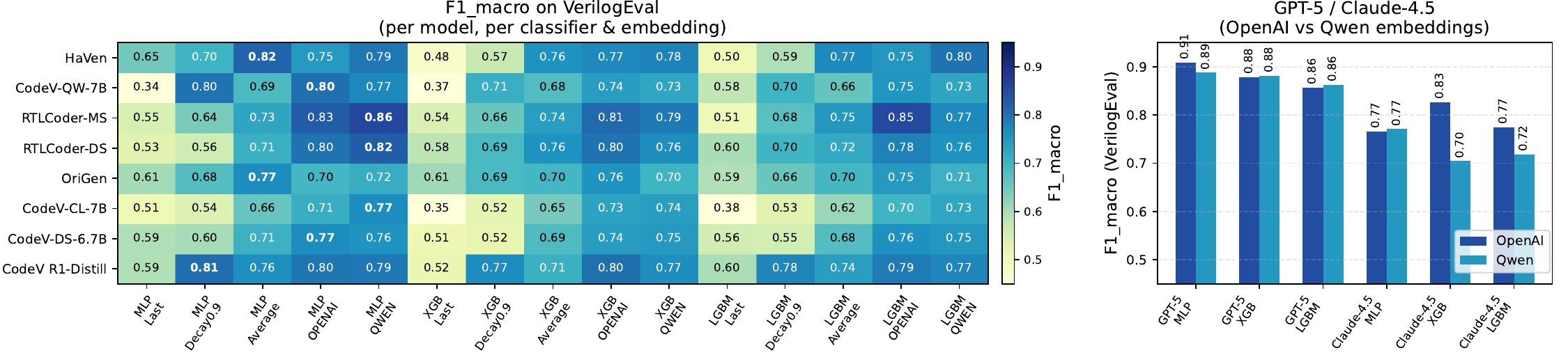} 
    \caption{Classifier F1-macro scores on VerilogEval benchmark comparing embedding strategies and classifier architectures.}
    \label{fig:verilogeval-classifier}
\end{figure*}
\begin{figure*}[!t]
    \centering
    \includegraphics[width=1.0\linewidth]{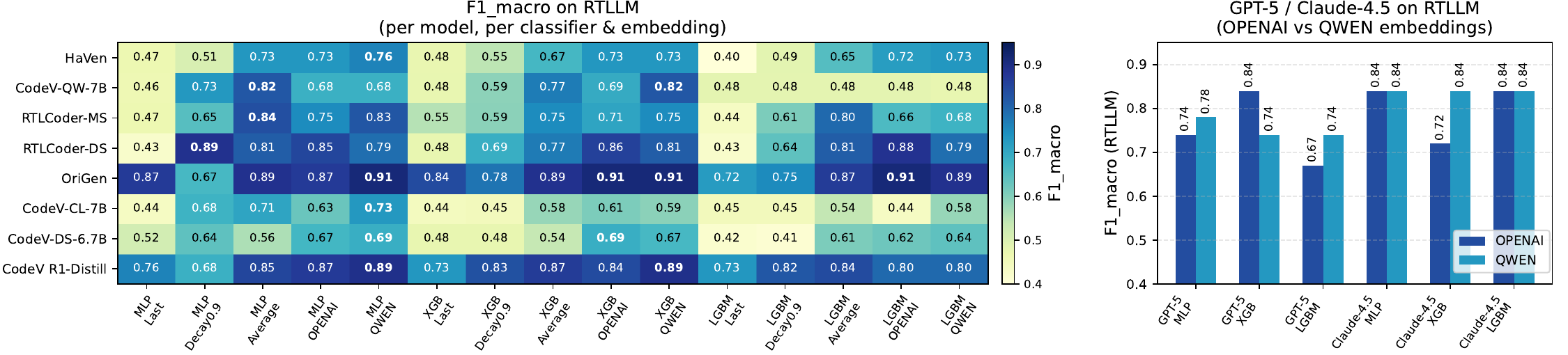} 
    \caption{Classifier F1-macro scores on RTLLM benchmark comparing embedding strategies and classifier architectures.}
    \label{fig:rtllm-classifier}
\end{figure*}

\subsection{Design Embedding}
LLMs possess self-awareness capabilities even before generation ~\cite{ni2025towards}, encoding their understanding of task complexity in hidden representations. We exploit this property by extracting semantic embeddings from the model's last layer to capture intrinsic difficulty perception.
For open-source models with accessible internals, we extract three complementary embedding types using attention masks from last layer: (1) \textbf{last token} embedding as the final compressed representation, (2) \textbf{average pooling} across all tokens for holistic semantics, and (3) \textbf{weighted decay pooling} for holistic but emphasizing recent context. Each view captures different aspects of model-perceived complexity (Fig.~\ref{fig:veridispatcher} (3)). For commercial models (e.g., GPT-5 and Claude-4.5) where internal hidden states are inaccessible, we use external embedding models, \texttt{text-embedding-3-large} and \texttt{Qwen3-Embedding-8B}, as semantic proxies. While these cannot capture model-specific architectural biases, they provide strong general-purpose semantic features enabling unified difficulty prediction across our heterogeneous model pool.

To probe perception boundaries and improve classifier generalization, we apply prompt-based augmentation to each benchmark task. As illustrated in Fig.~\ref{fig:veridispatcher} (1)(2), we generate multiple semantic variants per task by systematically varying linguistic formulation, technical vocabulary, abstraction level, and presentation style while preserving core functionality. 
This augmentation shows how well each model’s sense of task difficulty holds up across different ways of phrasing the same requirement, giving richer training data for the downstream difficulty predictor.

\subsection{Difficulty Classifier}
Defining task difficulty for Verilog generation requires comprehensive evaluation beyond syntax checking. Generated designs may appear structurally correct yet fail functionally. We therefore evaluate each model across three dimensions: syntax validity, structural similarity via Dolos~\cite{maertens2024discovering}, and functional equivalence checking.


\noindent
\textbf{Difficulty Score}: For each remaining augmented task variant, we perform 10 independent generations per model and aggregate the difficulty score $S_{diff}$ into a filtered difficulty score $filterd(S_{diff})$ in [0,1] representing the empirical failure rate (Fig.~\ref{fig:veridispatcher} (4)). During the filtration process~\cite{wang2025verireason}, we filter score based on the mean value falling within $[0.3, 1.8]$ and standard deviation exceeding 0.1, which removes samples that are either too trivial (consistently high success) or too difficult (consistently low success).Applying a threshold of 0.5, we create binary labels: tasks scoring above 0.5 are labeled \texttt{hard}, while those below 0.5 are \texttt{easy}. This model-specific labeling captures that difficulty varies across LLMs.

\noindent
\textbf{Training Classifiers}: Using the extracted embeddings and binary labels, we train lightweight classifiers, MLP, \textit{XGBoost} (\textit{XGB}~\cite{Chen16}), and \textit{LightGBM} (\textit{LGBM}~\cite{Ke17}), for each model. Once trained, these predictors enable pre-inference difficulty assessment(Fig.~\ref{fig:veridispatcher} (5)): given a new task description, each classifier predicts whether its corresponding LLM can successfully generate correct code without requiring actual generation or verification.

\subsection{Difficulty-Aware Dispatching}
With trained difficulty predictors, VeriDispatcher determines which model(s) to invoke for each task, balancing success rate and computational cost (Fig.~\ref{fig:veridispatcher} (6)). For each incoming task, the framework queries all model-specific classifiers using their optimal embedding type (determined via validation performance) to obtain difficulty predictions. Each classifier outputs a binary prediction indicating whether its corresponding model can successfully handle the task.
To accommodate deployment scenarios with varying cost and performance requirements, it provides two configurable mechanisms: \textbf{ (1) Commercial Model Inclusion}: it can include or exclude expensive commercial models (GPT-5, Claude-4.5) from the candidate pool, enabling users to trade cost for quality or maintain privacy through open-source-only routing. \textbf{(2) Top-k Selection}: Rather than dispatching to a single model, it also support top-k dispatching where k models with highest predicted calibrated success  probability are selected. 
For $k=1$, the task is assigned to the most confident model; for $k > 1$, multiple models attempt generation in parallel, increasing success rate at higher computational cost. By routing each task to models predicted to handle it well, VeriDispatcher leverages model specialization to achieve superior aggregate performance compared to single-model baselines or random selection.

\section{Experiment Setups}

\begin{table}[t]
\centering
\footnotesize
\setlength{\tabcolsep}{4pt}
\renewcommand{\arraystretch}{1.1}

\begin{tabular}{lcccccc}
\toprule
\multirow{2}{*}{\textbf{Method}} & 
\multicolumn{3}{c}{\textbf{VerilogEval (pass@k)}} & 
\multicolumn{3}{c}{\textbf{RTLLM (pass@k)}} \\
\cmidrule(lr){2-4} \cmidrule(lr){5-7}
 & \textbf{1} & \textbf{5} & \textbf{10} 
 & \textbf{1} & \textbf{5} & \textbf{10} \\
\midrule
HaVen                 & 0.48 & 0.56 & 0.60 & 0.26 & 0.42 & 0.42 \\
CodeV-QW-7B           & 0.45 & \textbf{\color{orange!80!black}0.65} & 0.69 & 0.04 & 0.28 & 0.40 \\
RTLCoder-Mistral      & 0.36 & 0.50	& 0.56 & 0.24 & 0.40 & 0.46 \\
RTLCoder-DeepSeek     & 0.26 & 0.51	& 0.58 & 0.28 & 0.36 & 0.40 \\
OriGen                & 0.33 & 0.56	& 0.63 & \textbf{\color{orange!80!black}0.34} & 0.42 & \textbf{\color{orange!80!black}0.48} \\
CodeV-CL-7B           & 0.31 & 0.56	& 0.62 & 0.18 & 0.34 & 0.44\\
CodeV-DS-6.7B         & 0.46 & 0.64	& \textbf{\color{orange!80!black}0.71} & 0.22 & 0.38 & 0.44 \\
CodeV-R1-Distill-Qwen-7B & \textbf{\color{orange!80!black}0.52}	& \textbf{\color{orange!80!black}0.65} & 0.70 & 0.32 & \textbf{\color{orange!80!black}0.46}	& 0.46 \\
\midrule
\addlinespace[-1pt]
\multicolumn{7}{l}{\textit{\scriptsize Commercial Models}}\\[-2pt]
GPT-5                 & 0.70 & 0.76	& 0.79 & 0.24 & 0.34 & 0.38 \\
Claude-4.5-Sonnet            & \textbf{\color{green!50!black}0.78} & \textbf{\color{green!50!black}0.84} & \textbf{\color{green!50!black}0.86} & \textbf{\color{green!50!black}0.42} & \textbf{\color{green!50!black}0.44} & \textbf{\color{green!50!black}0.44} \\
\midrule
\multicolumn{7}{l}{\textbf{Dispatching strategies using MLP}} \\
\textit{Average Emb.-wo}          & \textbf{\color{orange!80!black}0.43} & \textbf{\color{orange!80!black}0.62} & \textbf{\color{orange!80!black}0.69} & 0.26 & 0.44 & \textbf{\color{orange!80!black}0.50}{\color{orange!80!black}{$\uparrow$}}\\
\textit{Qwen Emb.-wo }              & \textbf{\color{orange!80!black}0.43} & 0.61 & 0.65 & 0.32 & 0.44 & 0.48 \\
\textit{OpenAI Emb.-wo }            & 0.38 & 0.60 & 0.65 & \textbf{\color{orange!80!black}0.36}{\color{orange!80!black}{$\uparrow$}} & \textbf{\color{orange!80!black}0.46} & \textbf{\color{orange!80!black}0.50}{\color{orange!80!black}{$\uparrow$}}  \\
\textit{Average+OpenAI Emb.-w }     & \textbf{\color{green!50!black}0.69} & 0.76 & 0.80 & 0.36	& 0.46 & \textbf{\color{green!50!black}0.50}{\color{green!50!black}{$\uparrow$}} \\
\textit{Qwen Emb.-w }               & 0.65 & 0.74 & 0.79 & 0.38	& \textbf{\color{green!50!black}0.48}{\color{green!50!black}{$\uparrow$}} & \textbf{\color{green!50!black}0.50}{\color{green!50!black}{$\uparrow$}} \\
\textit{OpenAI Emb.-w }             & \textbf{\color{green!50!black}0.69} & \textbf{\color{green!50!black}0.78} & \textbf{\color{green!50!black}0.79}  & \textbf{\color{green!50!black}0.38} & 0.46 & \textbf{\color{green!50!black}0.50}{\color{green!50!black}{$\uparrow$}} \\
\bottomrule
\end{tabular}

\caption{Pass@k results for VerilogEval and RTLLM. Commercial bests are in green; open-source bests in orange.}
\label{tab:passk_results}
\end{table}

\subsection{Dataset Preparation}

We evaluate VeriDispatcher on two standard RTL generation benchmarks: RTLLM~\cite{lu2024rtllm} (50 cases) and VerilogEval~\cite{liu2023verilogeval} (156 cases). To mitigate limited dataset size, we use Claude-4.5-Sonnet to reverse engineer each reference design into ten prompt variants. For each variant, we sample ten generations and verify them with simulation and synthesis. This process captures LLM stochasticity and yields rich difficulty signals.
The expanded datasets comprise total 1,716 tasks with 17,160 generations for VerilogEval, and total 550 tasks with 5,500 generations for RTLLM. The original benchmarks (156 VerilogEval and 50 RTLLM) are reserved for dispatching evaluation only, while the expanded variants are used for difficulty-aware training. Each task generation is labeled with verification outcomes and assigned an initial difficulty score $S_{diff}$. We aggregate each task's ten inference scores using metrics from ~\cite{wang2025verireason} to obtain a difficulty value $filterd(S_{diff})$. Tasks are labeled as \texttt{hard} (score $>$ 0.5) or \texttt{easy} (score $\leq$ 0.5).


\subsection{Model Selection}
We select a diverse set of LLMs based on three principles. \textbf{(1) Domain Specialization:} We include Verilog-specific fine-tuned models (HaVen-CodeQwen, RTLCoder-Mistral, RTLCoder-DeepSeek, OriGen, CodeV-QW-7B, CodeV-CL-7B, CodeV-DS-6.7B) to capture domain adaptation effects. \textbf{(2) Reasoning Diversity:} We include reasoning models (CodeV-R1-Distill-Qwen-7B, GPT-5, Claude-4.5-Sonnet), enabling difficulty encoding analysis  with intermediate reasoning chains. \textbf{(3) Model Source Balance:} We include open-source models and commercial models (GPT-5, Claude-4.5-Sonnet) to reflect real-world deployment diversity. For embedding extraction, we use three types: (i) the target model's own hidden states (using last token, average, or decay pooling with weight 0.9), (ii) \texttt{text-embedding-3-large}, and (iii) \texttt{Qwen3-Embedding-8B}, enabling cross-model compatibility for difficulty prediction evaluation.


\subsection{Predictor Training}
We evaluate three classifier families: MLP, XGB, and LGBM. MLPs model non-linear structure, while XGB and LGBM are well-suited for tabular embeddings with interpretability.We conduct grid search over the ranges: MLP hidden sizes \{(512, 256), (1024, 512)\}, dropout \{0.1, 0.3, 0.5\}, learning rates \{1e-3, 1e-4, 1e-5\}; XGB/LGBM learning rates \{0.05, 0.1\}, estimators \{500, 800\}, max depth \{20, 30\} for XGB and leaves \{64, 128\} for LGBM. Using an \texttt{80/20} train-validation split with early stopping, we identify the optimal configurations: MLP with hidden size \texttt{(512, 256)}, dropout \texttt{0.1}, learning rate \texttt{1e-4}; XGB/LGBM with learning rate \texttt{0.05}, \texttt{500} estimators (max depth \texttt{20} for XGB; \texttt{64} leaves for LGBM). As MLP achieves the best performance, we apply temperature scaling~\cite{guo2017calibration} on the validation set to improve probability calibration for reliable easy/hard predictions. Unless explicitly stated, all classification tasks are performed using an MLP classifier.

\section{Experiment Results}

\subsection{Embedding Approach Comparison}

Embedding strategy comparison reveals clear, dataset-dependent patterns. On VerilogEval (Fig.~\ref{fig:verilogeval-classifier}), external embeddings (OpenAI, Qwen) form a tight cluster, reaching up to 0.86 F1-macro for open-source models and 0.91 for commercial models, reflecting the benchmark's strong semantic separability. The minimal gap between encoders (0.01--0.04) suggests either encoder sufficiently captures VerilogEval's semantic structure among task variants. For internal embeddings, average pooling consistently outperforms last-token and decay pooling, indicating that token-level aggregation yields more stable semantic representations than relying on a single terminal token~\cite{jiang2024repurposing}. Notably, average-pooled internal embeddings can approach external embedding performance.

In contrast, RTLLM (Fig.~\ref{fig:rtllm-classifier}) exhibits far broader performance range (0.48--0.91) even for external embeddings, revealing more fragmented and irregular difficulty boundaries driven by richer control-flow and protocol semantics. This variance is amplified with internal embeddings, where performance differences across LLMs become more pronounced, several fall below 0.50 even with average pooling. However, models with stronger RTLLM understanding, such as OriGen (best performer in Tab.~\ref{tab:passk_results}), achieve strong embedding quality across all pooling strategies. These results suggest RTLLM requires more discriminative, globally trained representations to properly capture its structural and protocol-level complexity.

Overall, external embedding encoders provide consistent difficulty signals across varying task complexity, while internal encoders' effectiveness correlates with the base model's domain proficiency models excelling at RTL generation naturally produce more informative embeddings for difficulty prediction.

\begin{figure*}[!t]
    \centering
    \includegraphics[width=1.0\linewidth]{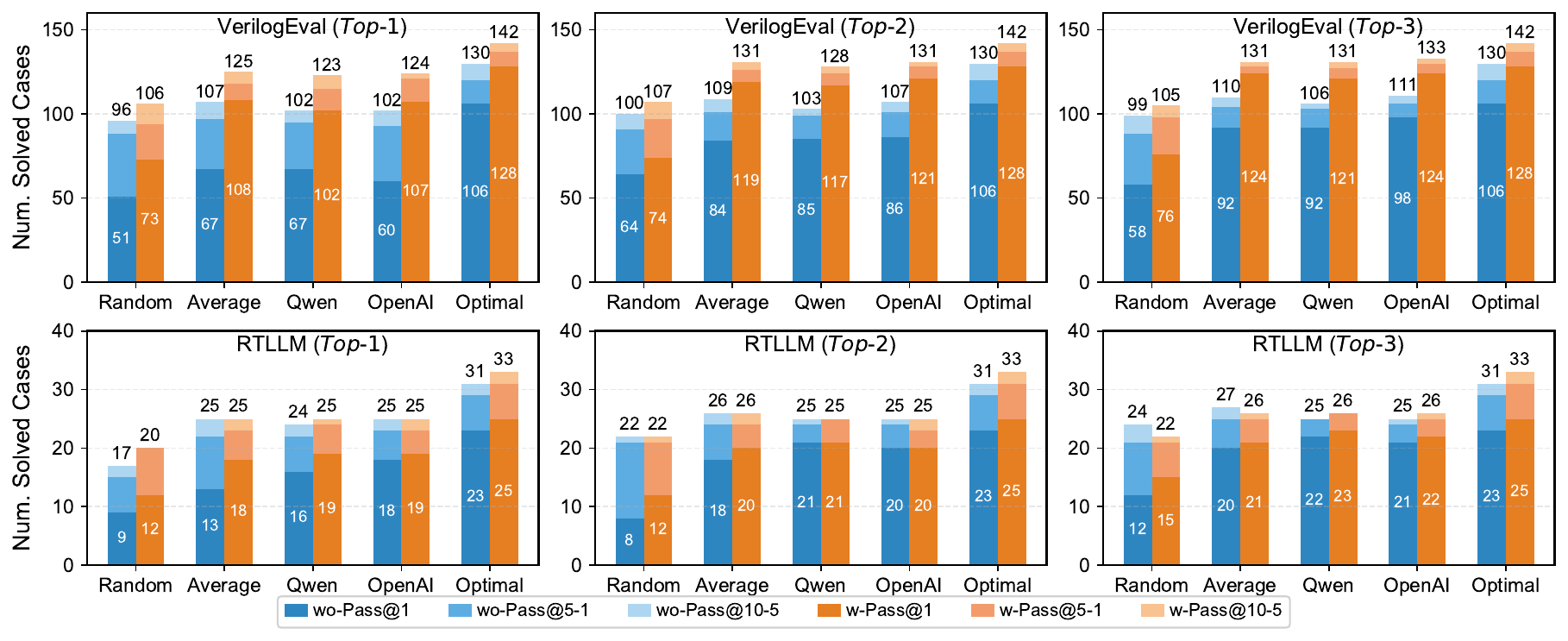} 
    \caption{Number of solved cases across Top-K dispatching strategies for VerilogEval and RTLLM benchmarks.}
    \label{fig:dispatching}
\end{figure*}

\subsection{Training Algorithms Analysis}

Classifier performance exhibits clear dataset-dependent behavior. On VerilogEval, top-performing models such as GPT-5 and Claude-4.5 Sonnet achieve nearly classifier-invariant accuracy (0.86--0.91; Fig.~\ref{fig:verilogeval-classifier}). This stability across MLP, XGBoost, and LightGBM shows that when embeddings already provide well-structured difficulty information, classifier choice has little impact. Because VerilogEval is semantically focused, even simple classifiers perform well with high quality embeddings, which indicates that is embedding quality, rather than architecture, is the primary factor.

RTLLM, however, shows a different pattern on Fig.~\ref{fig:rtllm-classifier}. While GPT-5 remains consistently stable ($\sim$0.84) across classifiers, Claude-4.5 Sonnet varies noticeably--strong with MLP and LightGBM but weaker with XGBoost. Open-source models exhibit even larger swings even up to 0.34 F1-macro, indicating that RTLLM's fragmented difficulty boundaries require more expressive models to capture non-linear decision regions. Tree-based methods behave inconsistently, suggesting the difficulty landscape does not align with axis-parallel splits, whereas MLPs generalize more reliably.

Cross-model trends support this distinction. Commercial models are stable across settings, while open-source models exhibit dataset and classifier dependent variation. HaVen degrades under certain classifier-embedding combinations, whereas OriGen improves under specific configurations. Overall, effective difficulty prediction requires joint optimization of embeddings and classifier architecture, especially on structurally complex benchmarks like RTLLM; in contrast, semantically driven tasks like VerilogEval rely primarily on embedding strength with minimal classifier sensitivity. This suggests that deployment strategies should be task-aware: lightweight classifiers suffice for semantically-focused benchmarks, while structurally complex tasks demand careful classifier-embedding co-design.

\subsection{Dispatching Performance}

Tab.~\ref{tab:passk_results} shows pass@k results of VeriDispatcher using MLP versus individual models. The results show distinct trends across the two benchmarks, confirming the effectiveness of difficulty-aware dispatching.

On VerilogEval, our best strategy using a combined averaging scheme with OpenAI embeddings achieves a pass@1 score of 0.69, nearly matching GPT-5's 0.70 while clearly outperforming all open-source models. With larger sampling budgets, it reaches 0.78 pass@5 and 0.79 pass@10 (matching GPT-5). This shows that model dispatching can replicate premium performance by combining diverse model strengths.Without commercial models, we still achieve 0.69 pass@10, surpassing six of eight open-source baselines. The advantage of OpenAI over Qwen embeddings, with a 0.04 absolute pass@1 gain, highlights the importance of high-quality task representations for difficulty prediction on semantically benchmarks.

Dispatching yields even greater gains on RTLLM, which involves complex, protocol-rich tasks. Our OpenAI-based configuration achieves 0.38 pass@1 and 0.50 pass@10, significantly outperforming GPT-5 (0.24 and 0.38), corresponding to 58\% and 32\% relative improvements. It also exceeds Claude-4.5-Sonnet at higher sampling rates. These results show VeriDispatcher delivers gains beyond any single model, especially under varying task difficulty. Notably, without OpenAI embeddings, our configuration reaches 0.36 pass@1 and 0.50 pass@10, outperforming most standalone models, including GPT-5. This demonstrates that VeriDispatcher remains practical under privacy constraints where external embedding services are unavailable. Overall, VeriDispatcher is a generalizable, scalable, and efficient routing solution for RTL generation.

\subsection{Dispatching Algorithm Comparison}

Fig.~\ref{fig:dispatching} compares \textit{Top-k} dispatching performance across three embedding types (average pooling, Qwen, OpenAI). Increasing $k$ improves solved counts by reducing dispatch noise and approximating the oracle case, where all models are evaluated and the best is chosen.

With OpenAI embeddings, \textit{Top-1} solves 102 cases (123 w/ commercial), \textit{Top-2} reaches 107 (131), and \textit{Top-3} hits 111 (133), recovering 85--94\% of optimal (130/142) using just three LLM calls. This trend holds across embeddings, suggesting pool size drives gains more than embedding choice. Improvements diminish with increasing $k$ (+4--5 cases from \textit{Top-1} to \textit{Top-2}, $\leq$3 cases to Top-3), with most gains achieved by $k$=2.
In contrast, RTLLM shows less $k$ sensitivity, with solved cases ranging 24--27 (without commercial) and 25--26 (with commercial). The primary gain occurs from \textit{Top-1} to \textit{Top-2} (38.46\% improvement for average pooling), after which performance plateaus, indicating difficulty structure saturates quickly.

Random selection performs substantially worse, solving $<$100 VerilogEval cases and $<$24 RTLLM cases—less than half of embedding-guided performance—confirming naive selection fails to exploit difficulty patterns. Even restricting to open-source models, average pooling \textit{Top-1} achieves 107 VerilogEval cases (31.37\% improvement over random), demonstrating strong dispatching effectiveness.
\subsection{Model Selection Analysis}
\begin{figure}[!t]
    \centering
    \includegraphics[width=1.0\linewidth]{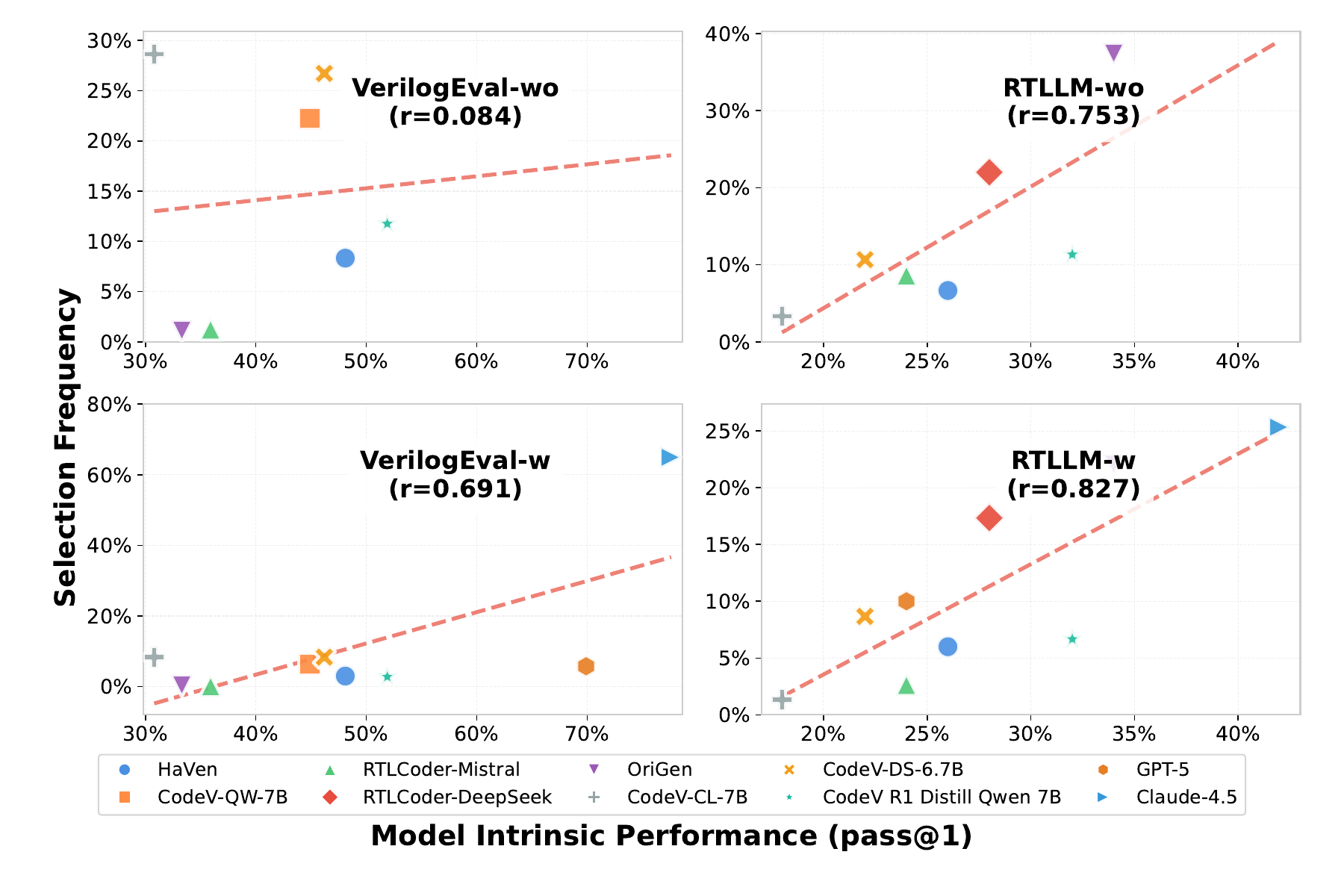} 
    \caption{Veridispatcher model selection statistics.}
    \label{fig:model_sel}
\end{figure}

We analyze the relationship between each model's intrinsic capability and its selection frequency within the VeriDispatcher framework. We evaluate both benchmarks with and without commercial models (Claude-4.5-Sonnet and GPT-5). To obtain stable selection trends, we aggregate results across all embedding variants, including average pooling, OpenAI, and Qwen embeddings.
As shown in Fig.~\ref{fig:model_sel}, for VerilogEval without commercial models (VerilogEval-wo), the correlation between model capability and selection frequency is weak (r = 0.084), consistent with the modest dispatching gains reported in Table~\ref{tab:passk_results}. When commercial models are included (VerilogEval-w), this correlation increases markedly (r = 0.691), suggesting that VeriDispatcher more reliably identifies strong performers, particularly Claude-4.5 -Sonnet, which is selected far more frequently.

In contrast, the RTLLM benchmark exhibits strong positive correlations in both configurations (r = 0.753 without commercial models and r = 0.827 with commercial models). This aligns with the substantial improvements observed in Table~\ref{tab:passk_results}, where VeriDispatcher surpasses the best single-model pass@10 performance by $\geq$2\% in both settings, demonstrating consistently effective model routing regardless of commercial model availability. 
This disparity reflects distinct routing dynamics. On RTLLM, complexity creates pronounced capability differentiation, models like OriGen excel on specific task types while others struggle, enabling effective dispatching. Conversely, VerilogEval's simpler semantics yield more uniform performance across models, leaving narrow capability gaps. Despite high prediction accuracy, marginal model differences lead to frequent routing misjudgments, particularly without strong commercial models to establish a clear performance ceiling.

As shown in Fig.~\ref{fig:dispatching}, \textit{Top-3} yields the best performance across all configurations; so we focus on its accuracy gains and commercial model usage. On VerilogEval, pass@10 with OpenAI's embeddings exceeds GPT-5 by 5.77\% while reducing commercial model usage by 25\%. On RTLLM, pass@10 with average pooling embedding improves performance by 12.5\% without any commercial model, and by 18.18\% with using only 40\% commercial models.

\subsection{VerilogEval: Constrains on classifier}

The performance gap between VeriDispatcher on RTLLM and VerilogEval largely stems from differences in semantic density. As shown in Table~\ref{tab:semantic_comparison}, RTLLM contain far richer descriptions with 363.90 tokens on average, nearly double VerilogEval's 183.28 tokens. This additional context directly strengthens classifier accuracy. Fig. ~\ref{fig:verilogeval-classifier} and ~\ref{fig:rtllm-classifier} show that RTLLM classifiers consistently reach 0.81--0.91 F1-macro, whereas VerilogEval tops out at 0.75--0.86, with a particularly large gap for open-source models. Richer task descriptions give difficulty predictors more semantic cues, allowing them to form cleaner decision boundaries and align tasks with the right models more reliably. In contrast, VerilogEval's concise prompts offer fewer discriminative features, limiting the predictor's ability to gauge difficulty and reducing the effectiveness of routing. This semantic bottleneck explains the difference in gains: on RTLLM, VeriDispatcher delivers a 58\% pass@1 improvement over GPT-5, driven by precise difficulty-aware routing. On VerilogEval, where semantic signals are weaker, improvements rely more on ensemble diversity than on accurate prediction.

\begin{table}[h]
\centering
\footnotesize
\caption{Semantic richness by average token number.}
\label{tab:semantic_comparison}
\begin{tabular}{cccc}
\toprule
\textbf{RTLLM (Qes.)} & \textbf{RTLLM (Ref.)} & \textbf{VerilogEval (Qes.)} & \textbf{VerilogEval (Ref.)} \\
\midrule
363.90 & 369.16 & 183.28 & 123.31 \\
\bottomrule
\vspace{-6mm}
\end{tabular}
\end{table}
\section{Conclusion} \label{sec:con}
VeriDispatcher optimizes RTL generation by intelligently dispatching tasks across multiple LLMs using pre-inference difficulty prediction. We train model-specific classifiers on semantic embeddings of task descriptions, enabling the system to match each task to the most capable model.
This approach significantly boosts generation quality, achieving up to 18.18\% higher on RTLLM and 5.77\% on VerilogEval compared to GPT-5, matching commercial models. VeriDispatcher remains effective under budget constraints, achieving strong results using only open-source models. Our analysis reveals that external embeddings and average pooling embedding outperform other internal states, semantic richness improves classifier accuracy, and task difficulty is model-dependent. VeriDispatcher provides a scalable and generalizable framework for real-world RTL generation, helping teams leverage diverse LLM portfolios while maintaining cost-efficiency and deployment flexibility.

Despite strong performance, VeriDispatcher is limited by the small size of current RTL benchmarks, which constrains difficulty prediction performance. 
Future work includes hierarchical dispatch from cheap to premium models, extending to other EDA tasks, e.g., testbench generation, assertion synthesis, and developing difficulty-aware embeddings.

\bibliographystyle{ACM-Reference-Format}
\bibliography{Reference}

\end{document}